\documentclass[runningheads]{llncs}
\usepackage[T1]{fontenc}
\usepackage{graphicx}
\usepackage{mathrsfs}
\usepackage{amsmath}
\usepackage{cite}
\usepackage{subfigure}
\usepackage{paralist}

\usepackage{listings}
\usepackage{algorithm}
\usepackage{algpseudocode}

\begin{document}
\title{Support Vector Based Anomaly Detection \\ in Federated Learning}
%
%
\author{Massimo Frasson\inst{1}\orcidID{0009-0008-1483-5021} \and
Dario Malchiodi\inst{1}\orcidID{0000-0002-7574-697X}}
\authorrunning{M. Frasson and D. Malchiodi}
%
\institute{Department of Computer Science, University of Milan,Via Celoria 18, 20133 Milan, Italy
\email{dario.malchiodi@unimi.it, massimo.frasson@studenti.unimi.it}}
\maketitle              
\begin{abstract}
Anomaly detection plays a crucial role in various domains, from cybersecurity to industrial systems. However, traditional centralized approaches often encounter challenges related to data privacy. In this context, Federated Learning emerges as a promising solution. This work introduces two innovative algorithms—Ensemble SVDD and Support Vector Election—that leverage Support Vector Machines for anomaly detection in a federated setting. In comparison with the Neural Networks typically used in within Federated Learning, these new algorithms emerge as potential alternatives, as they can operate effectively with small datasets and incur lower computational costs. The novel algorithms are tested in various distributed system configurations, yielding promising initial results that pave the way for further investigation.

\end{abstract}

\section{Introduction}
Machine Learning (ML) has become increasingly pervasive in contemporary life, manifesting in diverse applications within business and consumer products. A plethora of algorithms exist, tailored to specific tasks. Among these tasks, Anomaly Detection (AD) received particular attention during the past few decades~\cite{thudumu}.
Initially conceived to spot low-quality measurements in experimental outcomes, AD swiftly found broader applications due to the intrinsic value of anomalies, events characterized by unexpected and abrupt occurrences.  Anomalies, by their nature, are infrequent and often entail significant costs when they do transpire. Consider for instance continuous systems: being able to detect anomalous behaviors as they develop, for instance, within a piece of industrial equipment, or in the human body, enables engineers, in the first case, and physicians, in the second one, to act proactively.

By their nature, anomalies escape formal definitions (excluding, of course, that of not behaving as in normal conditions), therefore it is very difficult to find them by using traditional algorithmic methods. Indeed, AD is typically rooted in statistical techniques: notably, ML proved to be effective in detecting or foreseeing them~\cite{nassif2021machine}. However, the classical supervised approach in ML is not well suited to tackle the problem: this is because data describing anomalies are typically scarce, resulting in a strong class imbalance. This makes AD different from binary classification, and thus requiring different approaches. The more obvious option is that of unsupervised ML techniques: various algorithms have been proposed, as extensively detailed in \cite{10.1145/3381028,comparative_evaluation}. However, a significant drawback pervades the existing paradigm of AD when the objects to be analyzed live in a distributed world. Conventional algorithms require the centralization of all available data to derive a model able to distinguish between normal and anomalous behavior. Besides the obvious logistical challenges posed by the transmission and storage of substantial data volumes, this centralization introduces serious privacy concerns. The potential leakage of the collected data, either during transmission or as the effect of cyber attacks to the server, raises severe business issues, particularly concerning sensitive domains like those involving medical records, culminating in costly fines, such as the ones regulated by the GDPR~\cite{Voigt2017}.

Given the practical benefits of ML in our society, the imperative arises to derive equivalent value from data involving AD, yet removing privacy concerns from the equation. Enter Federated Learning (FL), a paradigm introduced by \cite{konevcny2016optimization,konevcny2016federated,chen2021communication}. FL is defined as a loose federation of devices (clients) participating in a ML process, under the coordination of a central server. FL adopts a protocol ensuring that the dataset of each client remains completely undisclosed, even to the server itself. Indeed, clients share with the central server only model updates computed via training on their local observations. The server simply receives these updates and aggregates them to build a global update which is subsequently distributed to all clients, so that they can enrich their performances without compromising privacy.
Within the FL domain, all challenges in the realm of distributed systems related to handling device federations over the Internet are addressed at a lower layer of abstraction. This abstraction is provided by the platform executing the decentralized learning process, enabling developers to focus on devising algorithms that perform comparably to their centralized counterparts within the federated framework.
One of the first studies in which FL was introduced~\cite{konevcny2016federated} also proposed the first federated algorithm, called \emph{FederatedAveraging}, devoted to training Neural Networks and applied to the problem of predicting the next word which will be typed using smart keyboards in mobile devices~\cite{hard2018federated}.

In general, Neural Networks are at the forefront of contemporary ML models, being able to capture complex and nonlinear relations among data. They have been extensively applied also to the AD domain (see, for instance, \cite{9925159} for a recent review). 
As highlighted in~\cite{PARK2023889}, the privacy concerns inherent in FL do not guarantee the availability of sufficient data to effectively train neural models on every single device. A trending solution in the literature is leveraging transfer learning. This involves training a general-purpose model on the central server and then fine-tuning it locally to provide a tailored model to each client, thereby protecting privacy and ensuring good personalized performance.
However, this solution still necessitates data availability on a central server, which is not always guaranteed in privacy-sensitive environments.
Training Neural Networks is computationally expensive, and while high-end smartphones currently available in the market are equipped with processors expressly designed for neural operations, they are more suited for querying complex networks, or fine-tuning them, rather than for a full training process~\cite{9767442}. This gap is further pronounced when considering the broader realm of IoT and edge devices~\cite{Thakur}.
%
%
These challenges leave room to explore alternative methodologies. Support Vector Machines (SVMs) have long been used as a valid computational counterpart w.r.t.\ Neural Networks, because of their ability to generalize without overfitting data, in the meanwhile possibly compressing the dataset onto a small set of \emph{support vectors} (SVs) that fully describe a learnt model. SVMs have been successfully applied to anomaly detection, modeling the region containing normal observations in two different ways: respectively, learning a hyperplane separating normal data and outliers~\cite{scholkopf2001estimating}, or building a small hypersphere enclosing normal observations~\cite{tax1999support}. The two approaches are equivalent under specific choices for the related hyper parameters, as shown in~\cite{scholkopf2001estimating}.

This work aims to integrate FL and the Support Vector Domain Description (SVDD) technique introduced in~\cite{tax1999support}, proposing two possible strategies for solving the AD problem in the federated context: Support Vector Election and Ensemble SVDD, sketched here below.
\begin{itemize}
\item
In the Support Vector Election approach, each client is tasked with the application of SVDD on its own data, subsequently applying a special noise addition mechanism inspired by differential privacy~\cite{dwork2006differential}, sampling in the vicinity of each SV in order to obtain a replacement which can be sent to the central server without privacy concerns. The server trains a new SVM on the accumulated support vectors, potentially discarding superfluous ones.
\item
The Ensemble SVDD approach also requires each client to apply SVDD on its local data, now transmitting the entire learned model (that is SVs, their weights, and the hyper parameter values) to the central server. Functioning as a relay, the server distributes the incoming models to all clients. The classification of new observations is done via ensemble voting, using all models. When at least one of them predicts that the observation is an inlier, it is classified as such. When all models detect an outlier, the observation is classified as anomalous. To ensure privacy, before transmitting the model, each client generates a synthetic dataset, and trains SVDD a second time on this dataset, to derive SVs that do not originate from the client's data.
\end{itemize}

Experimental assessments are conducted to evaluate the effectiveness of the proposed approaches and of the related noise addition mechanisms.
To maintain a comprehensive perspective and to facilitate meaningful comparisons, the experimental methodology aligns with that adopted in \cite{comparative_evaluation}, based on the OC-SVM approach~\cite{scholkopf2001estimating}, whose performances are used as a (centralized) baseline. By replicating and validating the related results, a seamless connection is established between the novel approaches presented here and the existing state-of-the-art methods.

This work is structured as follows: Sect.~\ref{sec:AD} describes the main approaches used in the literature to address the AD problem, while Sect.~\ref{sec:FL} is devoted to briefly describe the FL setting. Section~\ref{sec:proposed_methods} details the two proposed approaches for a federated-based anomaly detection, and Sect.~\ref{sec:experiments} illustrates the related experimental campaign, discussing the obtained results. Some concluding remarks end the paper.

\section{Anomaly Detection}\label{sec:AD}

Anomaly detection, also known as outlier detection, is a well-studied problem in ML and, more in general, in data mining. The goal of anomaly detection is to identify data points that behave in a significantly different way w.r.t.\ the distribution modeling the usual behavior of a system. Detecting such points is important to spot malfunctions (such as damages in a mechanical system, or bugs in software), to notice illicit use of resources (e.g., frauds in financial transactions of healthcare), or to get aware of the emergence of novel phenomena. 


One of the fundamental algorithms addressing AD
is Local Outlier Factor (LOF)~\cite{10.1145/342009.335388}: it is based on scoring a point $\mathbf{x}$ using the ratio of the average density of its $k$ nearest neighbors to the density of $\mathbf{x}$ itself. Points having a high LOF score are therefore considered local outliers. The effectiveness of this approach has been experimentally established, although its sensitivity to parameter selection and its high computational complexity have also been pointed out~\cite{comparative_evaluation}.
Analogous drawbacks are suffered by CBLOF~\cite{HE20031641}, originally introduced to outperform LOF and based on a more refined clustering procedure.
Statistical approaches, such as that adopted by HBOS~\cite{goldstein2012histogram}, evaluate anomalies by separately analyzing the various data features (typically via histogram analysis), subsequently aggregating the results into an outlier score. The obtained results outperform methods such as LOF, but this method assume that features are i.i.d., a property seldom satisfied by real-world datasets.

The most known adaptations of SVMs to AD are called  One-Class SVM (OC-SVM)~\cite{scholkopf2001estimating} and Support Vector Data Description (SVDD)~\cite{tax1999support}, both exhibiting advantageous results w.r.t.\ alternative approaches in scenarios involving sparse or complex datasets. The common approach in these algorithms consists in considering images of the data in a high-dimensional space $\mathscr{H}$, yet discriminating anomalies from normal data in two different ways: OC-SVM learns a hyperplane, whereas SVDD learns a sphere. In the setting considered here, given a set $\{ \mathbf{x}_1, \dots, \mathbf{x}_n \}$ of normal points, SVDD finds the multiplier values $\beta_1, \dots, \beta_n$ minimizing $\sum_{i, j}\beta_i \beta_j k(\mathbf{x}_j, \mathbf{x}_j)$, where $k$ denotes a Gaussian kernel, under the constraints $\sum_i \beta_i = 1$ and $0 \leq \beta_i \leq C \ \forall i = 1, \dots, n$. A new point $\mathbf{x}$ is classified in function of the distance between its image in $\mathscr{H}$ and the center of the learnt sphere, amounting to
\begin{equation}\label{rsquared}
R^2(\mathbf{x}) = 1 - 2 \sum_j \beta_j K\left(\mathbf{x}_j, \mathbf{x}\right) + \sum_{i, j} \beta_i \beta_j k\left(\mathbf{x}_i, \mathbf{x}_j\right) \enspace.
\end{equation}

It turns out that most of the optimal multipliers nullify, thus $R^2$ only depends on a small number of the original points, called support vectors. A crisp classification typically involves a comparison with the radius of the sphere, in turn obtained computing (\ref{rsquared}) on any support vector.

Despite their potential, the above-mentioned techniques for outlier detection may face practical challenges in distributed and privacy-sensitive settings. In many real-world scenarios, data is scattered across multiple sources or entities, and it may not be feasible or desirable to centralize it. The main contribution of this work (cfr.\ Sect.~\ref{sec:proposed_methods}) is to propose two extensions of the SVDD algorithm that are privacy-sensitive and distributed. Both of them are deeply rooted in the FL framework, introduced in the next section. 

\section{Federated Learning}\label{sec:FL}

The FL setting~\cite{chen2021communication} has been introduced to enhance the user experience in smartphones, using data collected via the latter to train ML models meanwhile preserving user privacy.  The core idea is to keep data on the devices and leverage their computing power to locally train a model. The involved devices can subsequently transmit their model updates to a central server. This server is in charge of aggregating these updates into a global model that is broadcast to all clients. This results, in principle, in a privacy-preserving learning methodology that gives each smartphone predictive capabilities obtained by exploiting the data of all users\footnote{For the sake of brevity, this description of the learning protocol is given in a simplified form. Some concern about the users privacy remain, as improperly crafted updates can leak information from the device, and this is handled in FL by adopting a \emph{secure aggregation} technique, together with encrypted communications and, possibly, differential privacy.}. FL is characterized by the following properties:
\begin{itemize}
    \item \emph{non-i.i.d.}: it is unlikely that data used in a local training procedure is representative of the entire user population;
    \item \emph{imbalance}: different users may exhibit varying behaviors and data generation patterns;
    \item \emph{massive distribution}: targeting mobile devices involves a huge number of clients;
    \item \emph{limited communication}: mobile devices may be offline, have limited data usage, or experience slow connections or low power charge.
\end{itemize}

Training within the FL framework typically takes place using \emph{FederatedAveraging}, an adaptation of Stochastic Gradient Descent that uses averaging to aggregate the weight updates of several neural networks having the same architecture, where each update has been locally learned by a single device using its own data batch. The average update is applied to the server's global model, repeating the process until a stopping condition is met. At that time, the global model is sent to all clients. The algorithm has been applied to the fields of image classification and language models~\cite{chen2021communication}, yielding impressive results.

The adoption of FL to perform AD has been predominantly done in the Internet of Things (IoT) domain, for instance in order to detect abnormal computation in complex systems behind augmented buildings~\cite{sater2021federated}, by considering the logs of indoor sensors and energy usage and focusing on an LSTM architecture. An analogous approach was applied to the identification of network attacks~\cite{9424138}, combining GRUs learnt from the single devices so as to obtain an ensemble voting system. Finally, \cite{nardi2022anomaly} addresses the issue of AD in FL under the assumption that clients might not agree on which the ``normal'' label is: after each client have executed OC-SVM on its data, pairs of clients exchange their models, and each uses the received model to classify its local data. In case of high accuracy, the two models agree about what is to be detected as normal: autoencoders can be trained separately on this data, subsequently applying \emph{FederatedAveraging}\footnote{Note that in this learning protocol there is a subtle violation of the FL paradigm: OC-SVM describes the learnt hyperplane using support vectors, that are points coming from the training set. The transmission of the model implies therefore the disclosure of some client data.}.

In general, the research around AD in FL is focused only on Neural Networks, leaving unexplored other fundamental model types. The next section proposes the direct use of SVMs in the FL framework.

\section{The proposed methods}\label{sec:proposed_methods}
This section describes the main contribution of this work: two federated learning algorithms for AD using SVMs. Replacing Neural Networks with SVMs removes the need to perform multiple rounds during training, thus these methods aim at attaining a better time complexity, possibly also improving prediction performances. The proposed algorithms are Ensemble SVDD (ESVDD), in which each client locally runs SVDD and the resulting models are used from the central server to obtain an ensemble learner, and Support Vector Election (SVE), whose clients select representative points via SVDD, to be sent to the server for a second-round learning phase, still using SVDD. The proposed algorithms are built upon the foundation established by~\cite{tax1999support}, though using most of the notation from~\cite{ben2001support}, which fits better the proposed learning algorithms.

\begin{algorithm}[t]
\caption{Ensemble SVDD}
\label{alg:esvdd}
\begin{algorithmic}
\Require $F$ (client fraction), $K$ (num.\ clients)
\State{$t \gets \max(F \cdot K, 1)$}
\State{$S \gets \text{selectRandomClients}(t)$}
\For{each client $k \in S$ \textbf{in parallel}}
 \State{$X \gets \text{retrieveData}(k)$}
 \State{$V, B, \gamma, C \gets \text{SVDD}(X)$}
 \State{$X' \gets \text{syntheticDataset}(V, B, X)$}
 \State{$m_k \gets \text{SVDD}(X')$}
\EndFor
\State{$M \gets \text{mergeModels}(m_1, \dots, m_t)$}
\State{$\text{sendToAllClients}(M)$}
\end{algorithmic}
\end{algorithm}

\subsection{Ensemble SVDD}
\label{cap3}
ESVDD recognizes that each client may have a distinct representation of the region containing normal points, due to local biases. Therefore, the algorithm uses an ensemble approach to leverage the collective insights of all clients, as detailed in Algorithm~\ref{alg:esvdd}. ESVDD initially selects a subset $S$ of clients for training, corresponding to a fraction $F$ of the total $K$ available clients. Each client $k$ independently trains a SVDD model on its local data $X$. The resulting model is described by the hyper parameter values ($C$ and $\gamma$), the support vector set $V$ and the multiplier set $B$. Note that outputting a prediction in the learnt model is related to the computation of (\ref{rsquared}), which involves all SVs, that is points from the training set. This means that clients cannot directly send their models to the central server without contradicting the FL principle prohibiting data from leaving the client's domain. To avoid this, ESVDD relies on a noise addition mechanism: after the local training, each client generates a synthetic dataset by drawing a sample from the input space and isolating the points within the learnt sphere\footnote{More precisely, sampling is based on a mixture of a normal distribution fitted to the dataset and several normal distributions of small variance, centered around SVs. The latter distributions are introduced to avoid that sampling be focused only on the densest regions of the dataset.}.
This process yields a new set of inliers on which SVDD is rerun, resulting in a new model with synthetic SVs not originating from the initial training set. This model can be therefore transmitted to the central server. The server redistributes the received models to all the clients (including the ones not participating in training) as an ensemble, which will classify a new point as normal if and only if it is predicted as such by at least one among the sent models.
As argued in \cite{comparative_evaluation}, it is better to address the prediction problem by assigning a score to the data points rather than a binary  ``inlier''/``outlier'' label, as some of the outliers might be more critical than others.
%
This is already done by each model in the ensemble. Thus the overall score is computed as follows:
if a point is predicted as an inlier, ESVDD returns the maximum score among all clients' score assignments; conversely, it returns the sum of scores from all clients. The latter strategy heavily penalizes points that deviate significantly from many clients.

\begin{algorithm}[t]
\caption{Support Vector Election}
\label{alg:sve}
\begin{algorithmic}
\Require $F$ (client fraction), $K$ (num.\ clients), $\gamma$ (kernel width), $C$ (slackness factor), $\sigma$ (standard dev.\ for anonymisation), $\tau$ (threshold for difference between distances)
\State{$t \gets \max(F \cdot K, 1)$}
\State{$S \gets$ $\text{selectRandomClients}(t)$}
\For{each client $k \in S$ \textbf{in parallel}}
    \State{$X \gets \text{retrieveData}(k)$}
    \State{$V, B \gets$ SVDD($X, \gamma, C$)}
    \State{$V'_k \gets \emptyset$}
    \For{$\mathbf{v} \in V$}
        \State{$\mathbf{q} \gets \text{gaussian}(\mathbf{v}, \sigma)$}
        \State{$R^2 \gets \text{distanceFromCenter}(V, B)$}
        \While{$|R^2(\mathbf{v}) - R^2(\mathbf{q})| > \tau$}
            \State{$\mathbf{q} \gets \mathbf{q} - \epsilon (\mathbf{q} - \mathbf{v})$}
        \EndWhile
        \State{$V'_k \gets V'_k \cup \{ \mathbf{q} \}$}
    \EndFor
\EndFor
\State{$V' \gets \cup_k V'_k$}
\State{$M  \gets  \text{SVDD}(V', \gamma, C)$}
\State{$\text{sendToAllClients}(M)$}
\end{algorithmic}
\end{algorithm}

\subsection{Support Vector Election}
\label{cap4}

Conceptually speaking, SVDD can be seen as an election process, selecting SVs as representatives for the entire dataset. Leveraging this observation, SVE adopts an election mechanism to select a small number of points transmitted to the central server, as detailed in Algorithm \ref{alg:sve}. Each client $k$ uses its local data $X$ to train a model via SVDD\footnote{Note that, in this case, the central server imposes uniform hyper parameter values ($C$ and $\gamma$) for all clients, and these values remain fixed throughout the algorithm.}. The resulting support vector set $V$ and multiplier set $B$ are then used to compute the function $R^2$ mapping points to the squared distance from the center of the learnt sphere in feature space. All SVs are then anonymised using a technique inspired by the Gaussian Mechanism \cite{TCS-042}: each support vector $\mathbf{v}$ is replaced by a new point $\mathbf{q}$ drawn from a Gaussian distribution centered around the vector itself. But in the case under study, this new point should also lie on the surface of the sphere learnt by SVDD. This problem is addressed by guiding sampled points towards the original support vector: if the initially drawn point lies too far from the surface, it is moved towards the original support vector, iterating the procedure until a satisfactory point is obtained. The result is a set $V'_k$ of anonymised support vectors, whose elements are transmitted to the central server, which merges them into a collective set $V'$. An SVDD run is subsequently run on this set, obtaining a model $M$ whose description is transmitted to all clients, which will use it in order to compute outlierness scores for new observations.

\begin{table}[t]
\caption{Performances of the considered algorithms, in terms of mean and standard deviation of AUC over several experiments with different $C$ values, over the considered datasets (first column, denoting for sake of brevity each dataset using its initial letter). Only the best and worst performing results over the various configurations adopted for SVE and SVDD, respectively marked with $+$ and $-$, are reported. Boldface highlights the configurations in which a federated approach is either in line with the centralized algorithms, or outperforms them.}
\label{tab:res_summary}
\setlength{\tabcolsep}{3.2pt}
\centering
\begin{tabular}
{ccccccc}
\hline
D & OC-SVM      & SVDD        & SVE ($+$)   & ESVDD ($+$) & SVE ($-$)   & ESVDD ($-$) \\ \hline
B & 0.98 ± 1E-3 & 0.93 ± 5E-3 & 0.95 ± 2E-2 & \textbf{0.96} ± 1E-2 & 0.63 ± 3E-1 & 0.68 ± 3E-1 \\
P & 0.94 ± 3E-2 & 0.79 ± 4E-4 & 0.80 ± 2E-2 & 0.88 ± 1E-2 & 0.48 ± 2E-1 & 0.60 ± 2E-1 \\
L & 0.59 ± 4E-3 & 0.56 ± 1E-4 & 0.57 ± 1E-2 & \textbf{0.79} ± 4E-3 & 0.48 ± 4E-2 & 0.51 ± 5E-2 \\
S & 0.91 ± 6E-3 & 0.77 ± 1E-4 & 0.79 ± 2E-2 & \textbf{0.88} ± 2E-3 & 0.53 ± 2E-1 & 0.62 ± 2E-1 \\
\hline
\end{tabular}
\end{table}

\section{Experiments and results}\label{sec:experiments}
Both ESVDD and SVE score the test data, with higher scores indicating a greater likelihood of the point being an inlier. This score is converted into a crisp classification via a threshold, and the evaluation of performances is done using the Area Under the Receiving Operator Curve (AUC). All experiments were conducted on the datasets Breast Cancer, Pen Global, Letter and Satellite introduced in \cite{comparative_evaluation}, employing min-max normalization to all features.

\subsection{Evaluating plasticity}
\label{subsec:plasticity}
The first part of the experiments concerns the ability of the proposed algorithms to be able to suitably adapt the learnt model to the provided dataset. In other words, the research question is: ``does FL preserve the ability of SVM methods for AD to avoid underfitting the data they process?''. As the experiments are not aimed at evaluating generalization ability (see Sect.~\ref{subsec:generalization} for this part), there is no separation between training and test set, meaning the whole dataset is used for training and evaluating a model.
To answer the research question, a comparison is made with two baselines:
\begin{inparaenum}[(i)]
\item OC-SVM and
\item SVDD
\end{inparaenum}
(both the algorithms are described in the Introduction).
Note that the considered baselines likely represent an upper bound to the performance achievable with federated techniques, as centralizing all data should provide them an unfair advantage w.r.t.\ FL. Following the methodology adopted in~\cite{comparative_evaluation}, the hyper parameter space is sampled, running each time the learning process and reporting the results in terms of mean and standard deviation of the AUC. The goal is to highlight the impact of the hyper parameter choice on these results. Concerning OC-SVM, the baseline is obtained sampling $\nu$ (a hyper parameter strictly related to $C$ in the modelisation adopted here) uniformly in $[0.2, 0.8]$ and employing an automatic tuning for $\gamma$. Due to limited computational resources, the experiments reported here only replicate the first hyper parameter sampling, fixing $\gamma$ at 1. The performances of SVDD, ESVDD, and SVE are replicated under the same setting. In all cases, $10$ hyper parameter samples were collected.
For computational constraints, ESVDD is run so that all client shared the values for $C$ and $\gamma$, and the two additional hyper parameters of SVE are fixed so that $\sigma=1$ and $\tau=10^{-3}$ (the latter serving as a measure of the surrogate's quality of anonymised support vectors). All the combinations of the total number of clients $K \in \{2, 5, 10\}$ and of the client fraction $F \in \{0.5, 1 \}$ are systematically tested. It is worth underlying that ESVDD and SVE are run both with and without using their anonymisation mechanisms, to discern the amount to which the latter impact in performance. Finally, the distribution of data w.r.t.\ clients is tested under two extreme scenarios: \emph{i.i.d.}, where the dataset is shuffled and evenly partitioned among clients, and \emph{Biased}, where observations are distributed across clients on the basis of KMeans clustering.
\begin{figure}[t]
\centering
\subfigure[SVE]{
\includegraphics[width=0.49\textwidth]{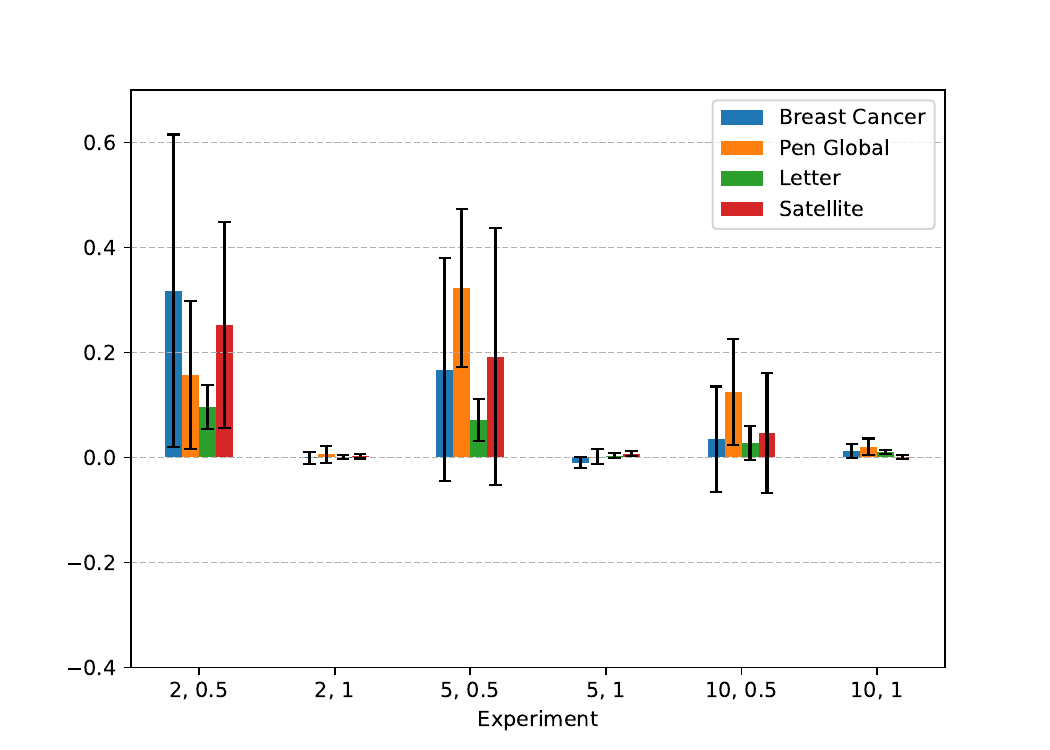}\label{fig:sve_splits}}
\subfigure[ESVDD]{\includegraphics[width=0.49\textwidth]{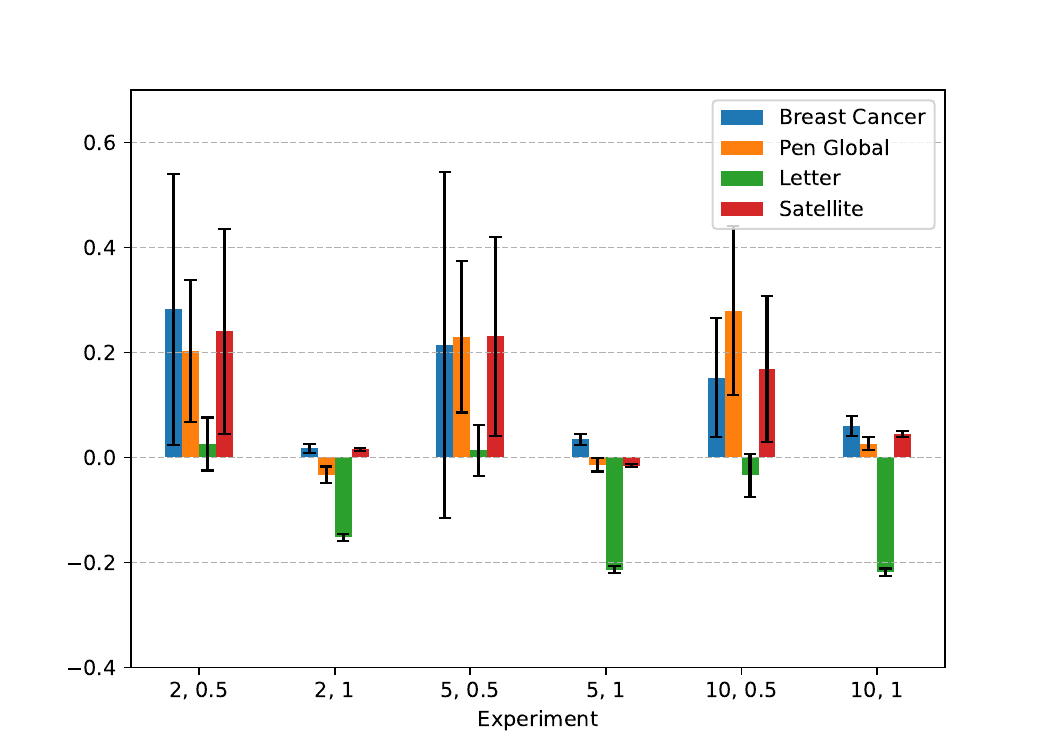}}
\caption{Impact on the performance of SVE (a) and ESVDD (b) of the data split distributions for different combinations of $K$ and $F$, shown in the X axis. The chart shows mean and standard deviation on several runs of the AUC difference, with positive values favoring i.i.d. over biased splits, and vice versa.
}
\label{fig:bias}
\end{figure}
Table~\ref{tab:res_summary} reports the obtained results. It is evident that SVE can match the performance of the baseline under optimal conditions (highlighted in bold in the table), although its performances w.r.t.\ the centralized counterparts significantly decrease under less favorable conditions. Similar considerations apply to ESVDD, with a remarkable difference: unexpectedly, this algorithm even outperforms the baseline under optimal conditions. Recall that the analyzed conditions refer to the following variables: degree of bias in the data distribution, client fraction, and total number of clients.
Figures~\ref{fig:bias} and~\ref{fig:fraction} analyze the configuration scenarios more in depth. More precisely, Fig.~\ref{fig:bias} depicts the AUC difference between experiments conducted in the \emph{i.i.d.} and \emph{Biased} settings, in function of the possible combinations of $K$ and $F$. In general, the difference is negligible when the client fraction is set to 1, but the impact becomes substantial when $F = 0.5$. This behavior is expected because with a client fraction of 0.5, only half of the clients are used in training: this is not problematic with i.i.d.\ data, whereas uncertainty and mismodeling can significantly affect performance when the data is strongly biased. However, it is worth nothing an interesting exception arising on the letter dataset: when $F=1$ the biased experiments always outperform those with i.i.d.\ data, and when $F=0.5$ the difference between the two settings is almost negligible.
\begin{figure}[t]
\centering
\subfigure[SVE]{\includegraphics[width=0.49\textwidth]{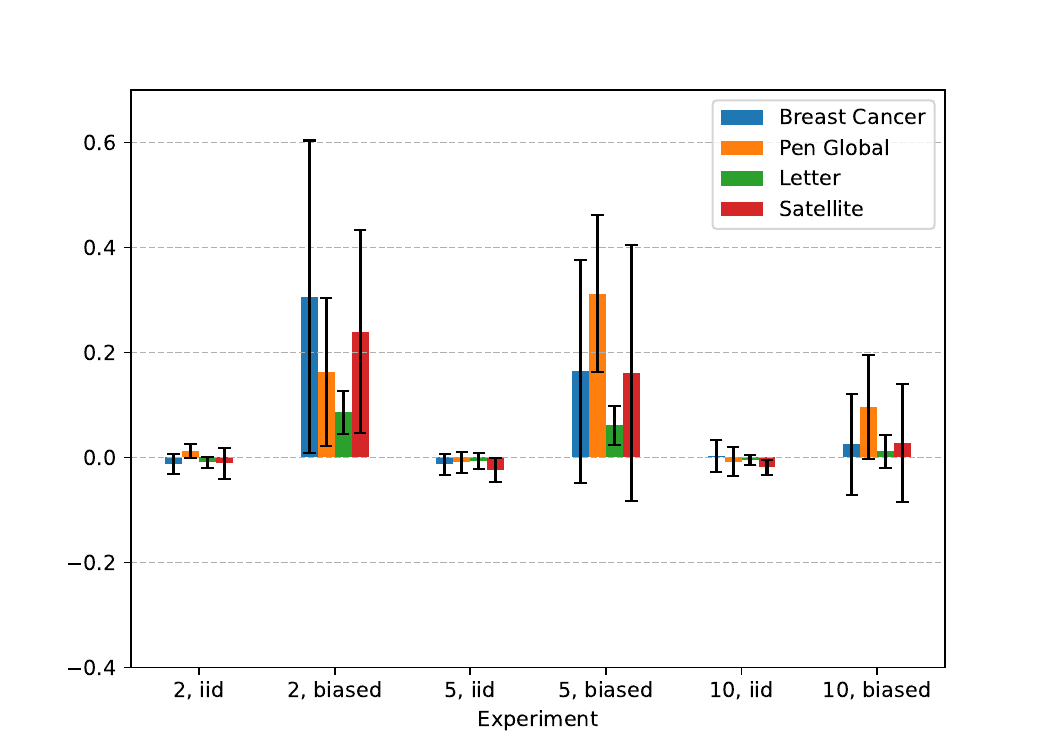}\label{fig:sve_frac}}
\subfigure[ESVDD]{\includegraphics[width=0.49\textwidth]{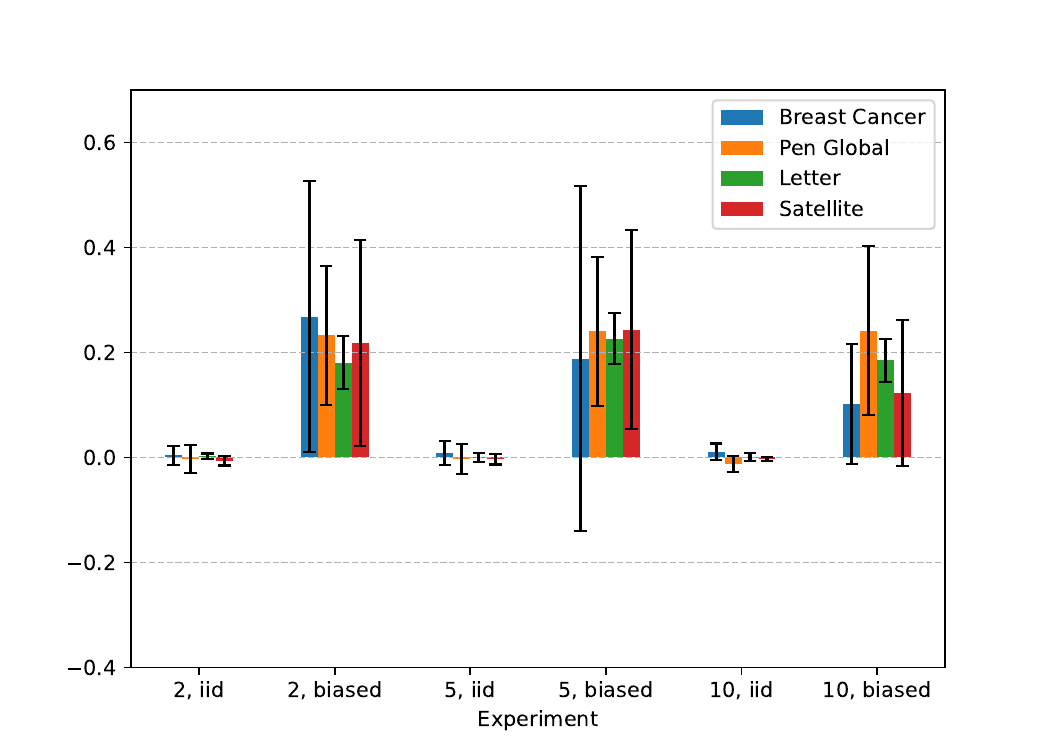}}
\caption{Impact on the performance of SVE (a) and ESVDD (b) of the client fraction $F$, for different combinations of $K$ and of the data split type. Same notation as in Fig.~\ref{fig:bias}, with positive values favoring the usage of all data over a subset, and vice versa.
}
\label{fig:fraction}
\end{figure}
Analogously, Fig.~\ref{fig:fraction} illustrates the impact of the client fraction when the client amount and bias are fixed. Also in this case, we observe as expected that:
\begin{inparaenum}[(i)]
\item i.i.d.\ data do not result in meaningful differences in performances, but a biased split have a considerable impact on the reported AUC values, and
\item this impact tends to be less evident as $F$ grows, especially when using SVE.
\end{inparaenum}
It is intriguing to highlight that, also in this experiment, the Letter dataset behaves in a different way w.r.t.\ the remaining datasets: using only a subset of the data leads to a result which is not dramatically worse than using all data, though this fact happens now in conjunction with SVE.
\begin{figure}
\centering
\subfigure[SVE]{\includegraphics[width=0.49\textwidth]{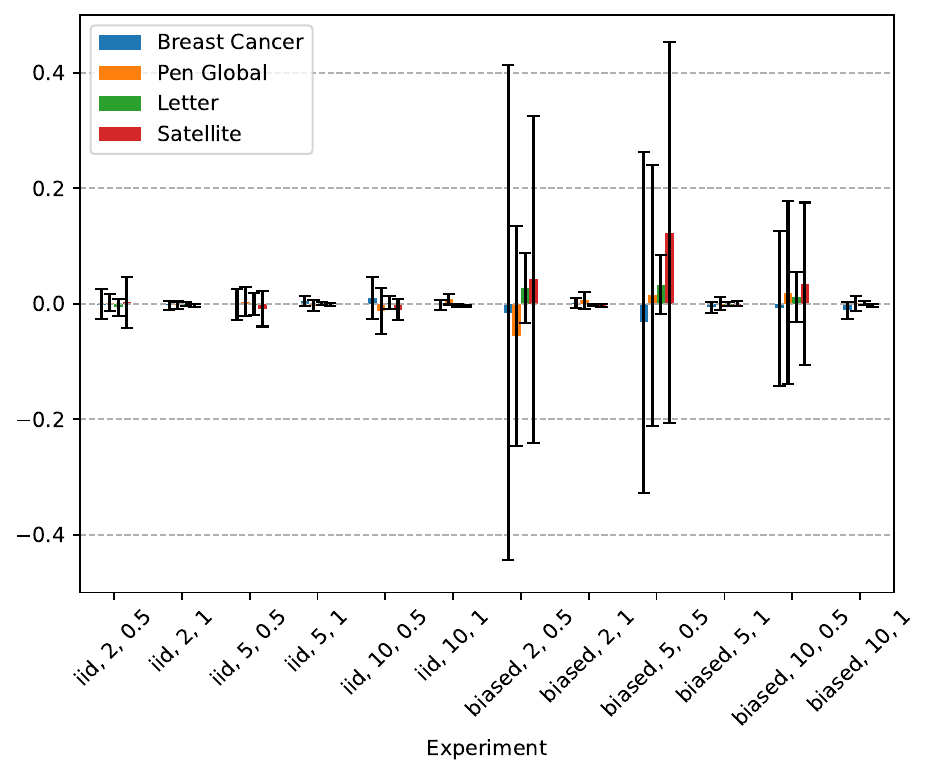}}
\subfigure[ESVDD]{\includegraphics[width=0.49\textwidth]{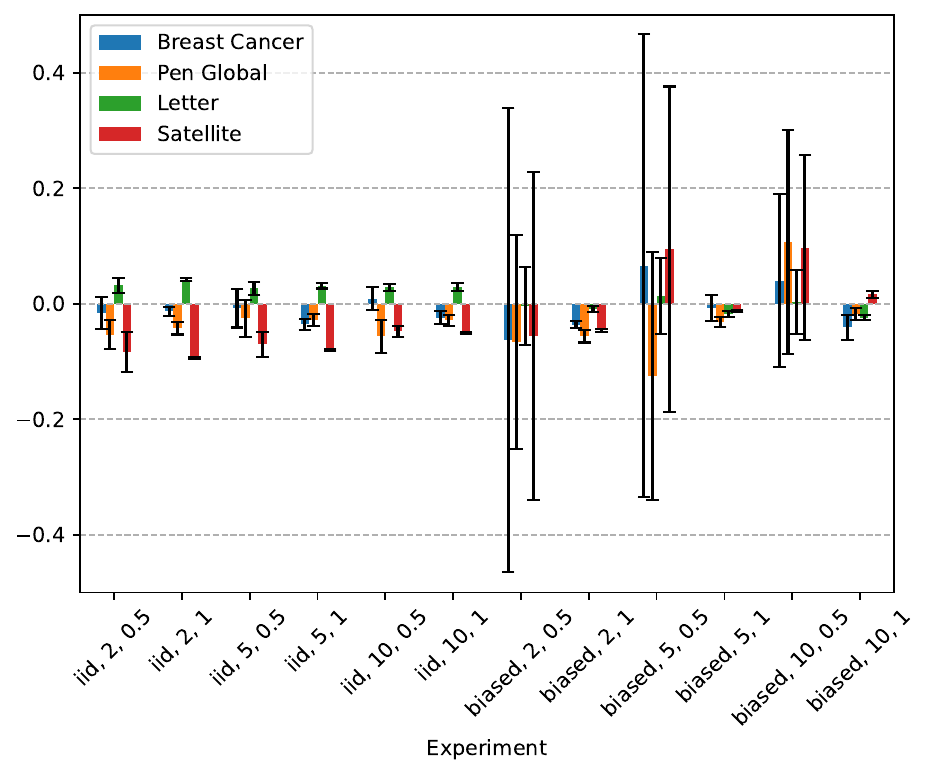}}
\caption{Impact of the anonymisation technique on the performance of SVE (a) and ESVDD (d), for each combination of $K$, $F$ and split bias. Same setting as in the previous figures, with positive values favoring using original data over applying anonymisation.}
\label{fig:noise_sve}
\end{figure}
Figure~\ref{fig:noise_sve} illustrates the impact of the anonymisation technique. It is evident that the use of anonymisation has minimal to no impact in most cases. However, in the likely worst-case scenario (i.e., $F=0.5$ and biased split) the difference in performance becomes more pronounced and uncertain. Rather than to anonymisation, this is likely due to the random client selection during training: indeed, experiments with and without noise may have selected different training clients, resulting in performance disparities. Overall, the impact of anonymisation remains relatively constant, although, unexpectedly, anonymisation tends to lead to better results. Finally, the experiments don't identify any discernible pattern regarding the impact of the number $K$ of clients.

\subsection{Evaluating generalization ability}
\label{subsec:generalization}
In order to asses the generalization ability of the proposed algorithms, a second set of experiments is conducted for each combination of the total number of clients $K \in \{2, 5, 10\}$ and the client fraction $F \in \{0.5, 1\}$ using a nested cross-validation process, with $3$ outer folds used to estimate the test error and $3$ inner folds devoted to the model selection phase.
More precisely, the best value for $C$ is found using a randomized search, uniformly sampling $10$ values in the range $[0.2, 0.8]$. The results are shown in Table~\ref{tab:res_cv_summary}, using the same notations as in Table~\ref{tab:res_summary} (although now the standard deviation refers to the different folds of the outer cross-validation).
The performances are in line with those shown in the previous section.
Instances in which the AUC values in Table \ref{tab:res_cv_summary} (highlighted in bold) surpass those in Table \ref{tab:res_summary} are frequent. This trend is rational since the former set of experiments involved averaging over multiple hyper parameter choices, some of which might have been sub-optimal. In the latter experiments, $C$ is optimized for each training set, resulting in better performances.
Of particular interest are the unexpected results for breast cancer, which exhibit markedly improved performances compared to the previous experiments. This suggests that the initial experiments might have been disadvantaged due to unfortunate choices of $C$ and the clients participating in the training.
In conclusion, the proposed methods, especially ESVDD, have the ability to generalize to unseen data.

\section{Conclusions}

The objective of this study is to adapt SV-based learning algorithms for AD to the FL context, proposing two novel algorithms named ESVDD and SVE. In both cases, the privacy requirements set by FL are addressed using special anonymisation techniques. A first comparison with state-of-the-art (that is, centralized) methods is conducted focusing on the total number of clients available for training, the fraction of clients actually participating in the learning process, and the degree of bias among client data. While centralized approaches generally exhibit the highest performance across datasets, the FL results are closest to this baseline when mimicking a centralized setting, i.e., all clients are used and data are split in an i.i.d.\ fashion. A trade-off emerges between split bias and fraction of clients, indicating that higher client fractions are needed for optimal performance in high bias scenarios. In some instances, the decentralized approach of FL even outperforms the centralized algorithms; this could be attributed to hyper parameter choices: ESVDD may be less affected by the adopted kernel width, and the use of $C$ in SVE differs from the centralized version, potentially justifying these differences in results. While no clear winner emerges between the two proposed algorithms, evidence suggests that the anonymisation technique performs better on SVE. In conclusion, while both proposed approaches are valid federated algorithms for AD, SVE stands out as a preferable choice, also in view of the fact that ESVDD increases its model size proportionally to the number of participating clients, making it impractical for massively distributed settings.
\begin{table}[t]
\caption{Mean and standard deviation over outer cross-validation folds of the AUC values of the experiments described in Sect.~\ref{subsec:generalization}, using the same notations as in Table~\ref{tab:res_summary}. Bold values highlight better results w.r.t.\ Table~\ref{tab:res_summary}.}
\label{tab:res_cv_summary}
\setlength{\tabcolsep}{2.5pt}
\centering
\begin{tabular}
{ccccccc}
\hline
D & OC-SVM & SVDD & SVE ($+$) & ESVDD ($+$) & SVE ($-$) & ESVDD ($-$) \\ \hline
B & 0.98 ± 7E-3 & 0.96 ± 3E-2 & \textbf{0.99} ± 5E-3 & \textbf{0.98} ± 9E-3 & \textbf{0.71} ± 2E-1 & \textbf{0.86} ± 1E-1 \\
P & 0.96 ± 1E-2 & 0.79 ± 6E-2 & \textbf{0.84} ± 8E-3 & 0.88 ± 1E-2 & \textbf{0.55} ± 2E-1 & 0.50 ± 3E-2 \\
L & 0.59 ± 4E-2 & 0.57 ± 3E-2 & \textbf{0.59} ± 2E-2 & 0.78 ± 2E-2 & \textbf{0.51} ± 5E-2 & \textbf{0.54} ± 2E-2 \\
S & 0.92 ± 8E-3 & 0.78 ± 8E-2 & \textbf{0.82} ± 4E-2 & \textbf{0.89} ± 3E-2 & \textbf{0.56} ± 2E-1 & \textbf{0.68} ± 2E-1 \\
\hline
\end{tabular}
\end{table}
Further experiments are crucial to confirm and extend these preliminary results. The algorithms should be tested on additional datasets, using a more realistic data split design and a sensibly higher number of clients, to better fit the FL context. In addition to that, the experiments done so far only perform a simple hyper parameter optimization: allowing the kernel width to be fine-tuned might enhance the overall performance, especially in high-bias settings; analogously, ESVDD could benefit from allowing each client to freely choose its hyper parameters, meanwhile filtering out or merging the models it receives in order to bar its inherent space complexity. Moreover, SVE might entail several training rounds, as in~\cite{chen2021communication}, in which clients collectively collaborate in finding the best hyper parameters. Finally, from a theoretical standpoint SVE lends itself to a formal Differential Privacy proof, a necessary step for comparing it with other privacy-preserving algorithms.

%
%
%
\bibliographystyle{splncs04}
\bibliography{mybibliography}
%




\end{document}